\documentclass[11pt,a4paper]{article}
\usepackage[hyperref]{emnlp-ijcnlp-2019}
\usepackage{times}
\usepackage{latexsym}
\usepackage{graphicx}
\usepackage{url}
\usepackage{amsmath}
\usepackage{booktabs}
\usepackage{cleveref}

\aclfinalcopy 

\title{Scalable and Accurate Dialogue State Tracking via Hierarchical Sequence Generation}

\author{Liliang Ren, Jianmo Ni and Julian McAuley \\
 University of California, San Diego \\
  La Jolla, CA92093 \\
   {\tt \{lren,jin018,jmcauley\}@ucsd.edu} \\}
  
\date{}

\begin{document}
\maketitle
\begin{abstract}
Existing approaches to dialogue state tracking rely on pre-defined ontologies consisting of a set of all possible slot types and values. Though such approaches exhibit promising performance on single-domain benchmarks, they suffer from computational complexity that increases proportionally to the number of pre-defined slots that need tracking. This issue becomes more severe when it comes to multi-domain dialogues which include larger numbers of slots. In this paper, we investigate how to approach DST using a generation framework without the pre-defined ontology list. Given each turn of user utterance and system response, we directly generate a sequence of belief states by applying a hierarchical encoder-decoder structure. In this way, the computational complexity of our model will be a constant regardless of the number of pre-defined slots. Experiments on both the multi-domain and the single domain dialogue state tracking dataset show that our model not only scales easily with the increasing number of pre-defined domains and slots but also reaches the state-of-the-art performance.

  

\end{abstract}

\section{Introduction}

\begin{table}[t!]
\begin{center}
\begin{tabular}{cc}
\toprule \bf DST Models & \bf ITC \\ \midrule
NBT-CNN \cite{mrkvsic2017neural} & $O(mn)$\\
MD-DST \cite{rastogi2017scalable}& $O(n)$  \\
GLAD \cite{zhong2018global}  & $O(mn)$ \\
StateNet\_PSI \cite{ren2018emnlp} & $O(n)$ \\
TRADE \cite{trade} & $O(n)$ \\
HyST \cite{HyST} & $O(n)$ \\
DSTRead \cite{dstread} & $O(n)$ \\
\bottomrule
\end{tabular}
\end{center}
\caption{\label{cpt} The Inference Time Complexity (ITC) of
previous DST models. The ITC is calculated based on how many times 
inference must be performed
to complete a prediction of the belief state in a dialogue turn, where $m$ 
is
the number of
values in a pre-defined ontology list and $n$ 
is
the number of slots.}
\end{table} 

A Dialogue State Tracker (DST) is a core component of a modular task-oriented dialogue system \cite{young2013pomdp}. For each dialogue turn, a DST module takes a user utterance and the dialogue history as input, and outputs a belief 
estimate
of the dialogue state. Then a machine action is decided based on the dialogue state according to a dialogue policy module, after which a machine response is generated.

Traditionally, a dialogue state 
consists of a set of \emph{requests} and \emph{joint goals}, 
both of which
are represented by a set of \emph{slot-value} pairs (e.g.~(\emph{request}, phone), (\emph{area}, north), (\emph{food}, Japanese)) \cite{henderson-thomson-williams:2014:W14-43}. 
In a recently proposed multi-domain dialogue state tracking dataset, MultiWoZ \cite{mwoz}, a representation of dialogue state consists of a hierarchical structure of domain, slot, and value is proposed. This is a more practical scenario since dialogues often include multiple domains simultaneously.

Many
recently proposed DSTs \cite{zhong2018global,ramadan2018large} 
are based on 
pre-defined ontology 
lists
that specify
all possible slot values in advance. To generate a distribution over the candidate set, previous works often take each of the \emph{slot-value} pairs as input for scoring. However, in 
real-world
scenarios,
it is
often
not practical to enumerate all possible slot value pairs and 
perform
scoring from a large dynamically changing knowledge base \cite{xu2018acl}. 
To tackle this problem, a popular direction is to build a fixed-length candidate set that is dynamically updated throughout the dialogue development.
\Cref{cpt} briefly summaries the inference time complexity of multiple state-of-the-art DST models following this direction. Since the inference complexity of all of 
previous model is at least proportional to the number of the slots, these models will 
\textit{struggle} to scale to
multi-domain datasets with much larger numbers of pre-defined slots.

\textbf{In this work}, we formulate the dialogue state tracking task as a sequence generation problem, instead of formulating the task as a pair-wise prediction problem as in existing work. We propose the \textbf{CO}nditional \textbf{ME}mory \textbf{R}elation Network (\textbf{COMER}), a scalable and accurate dialogue state tracker that has a constant inference time complexity. \footnote{The code is released at \url{ https://github.com/renll/ComerNet}}

Specifically, our model consists of an encoder-decoder network with a hierarchically stacked decoder 
to first generate the slot sequences in the belief state and then for each slot generate the corresponding value sequences. The parameters are shared among all of our decoders for the scalability of the depth of the hierarchical structure of the belief states. COMER applies BERT contextualized word embeddings \cite{bert} and BPE \cite{bpe} for sequence encoding to ensure the uniqueness of the representations of the unseen words. The word embeddings for sequence generation are initialized with the static word embeddings generated from BERT 
.

\section{Motivation}

\begin{figure}[t]
\centering
\includegraphics[width=\columnwidth]{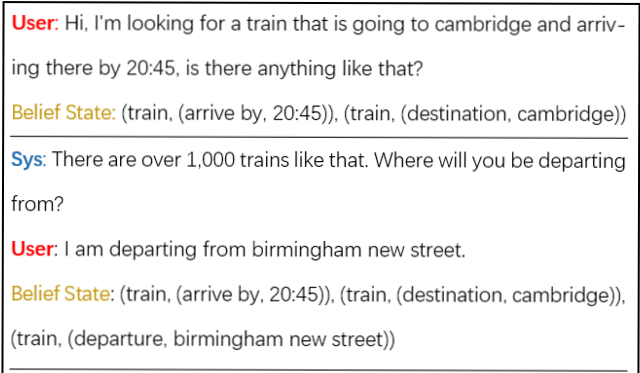}
\caption{An example dialogue from the multi-domain dataset, MultiWOZ. At each turn, the DST needs to output the belief state, a nested tuple of (DOMAIN, (SLOT, VALUE)), 
immediately
after the user utterance ends. The belief state is accumulated as the dialogue proceeds. Turns are separated by black lines.}\label{f1}
\end{figure}

\Cref{f1} shows a multi-domain dialogue 
in which
the user wants the system to first help book a train and then reserve 
a hotel.
For each turn, the DST will need to track the \emph{slot-value} pairs (e.g.~(arrive by, 20:45)) representing the user goals as well as 
the domain that the \emph{slot-value} pairs belongs to (e.g.~train, hotel).
Instead of representing the belief state via a hierarchical structure, one can also combine the domain and slot together to form a \emph{combined slot-value} pair (e.g.~(train; arrive by, 20:45) where the \emph{combined slot} is ``train; arrive by"), which ignores the subordination relationship between the domain and the slots.

\begin{figure}[t]
\centering
\includegraphics[width=\columnwidth]{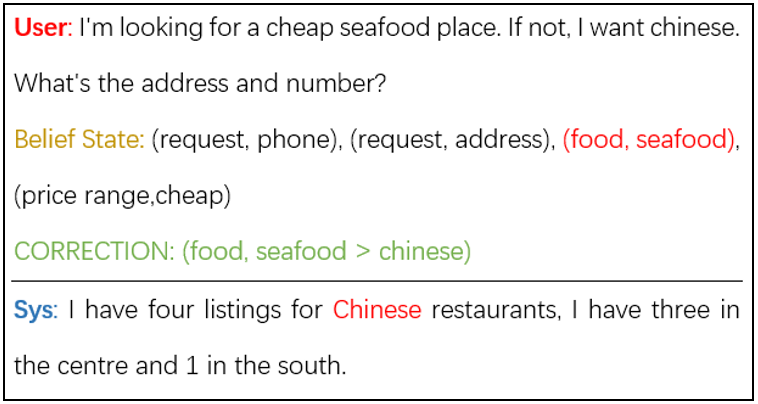}
\caption{An example 
in the WoZ2.0 dataset 
that invalidates the
\emph{single value assumption}. It is impossible for the system to generate the sample response about the Chinese restaurant with the original belief state (food, seafood). A correction 
could
be made as (food, seafood $>$ chinese) which has multiple values and a logical operator ``$>$".}\label{f2}
\end{figure}

A typical fallacy in 
dialogue state tracking 
datasets
is that they make an assumption that the slot in a belief state can only be mapped to 
a
single value in a dialogue turn. We call this the \emph{single value assumption}. Figure~\ref{f2} shows an example 
of this fallacy
from the WoZ2.0 dataset: Based on the belief state label (food, seafood), it will be impossible for the downstream module in the dialogue system to generate 
sample responses that 
return
information about 
Chinese restaurants. A correct representation of the belief state
could
be (food, seafood $>$ chinese). This
would
tell the system to first search the database for
information about 
seafood 
and then 
Chinese restaurants. The logical operator ``$>$" indicates which retrieved information should have a higher priority to be returned to 
the 
user. Thus we are interested in building DST modules capable of generating structured sequences, since this kind of sequence representation of the value is 
critical
for accurately capturing the belief states of a dialogue.


\begin{figure*}[htb]
\centering
\includegraphics[width=16cm]{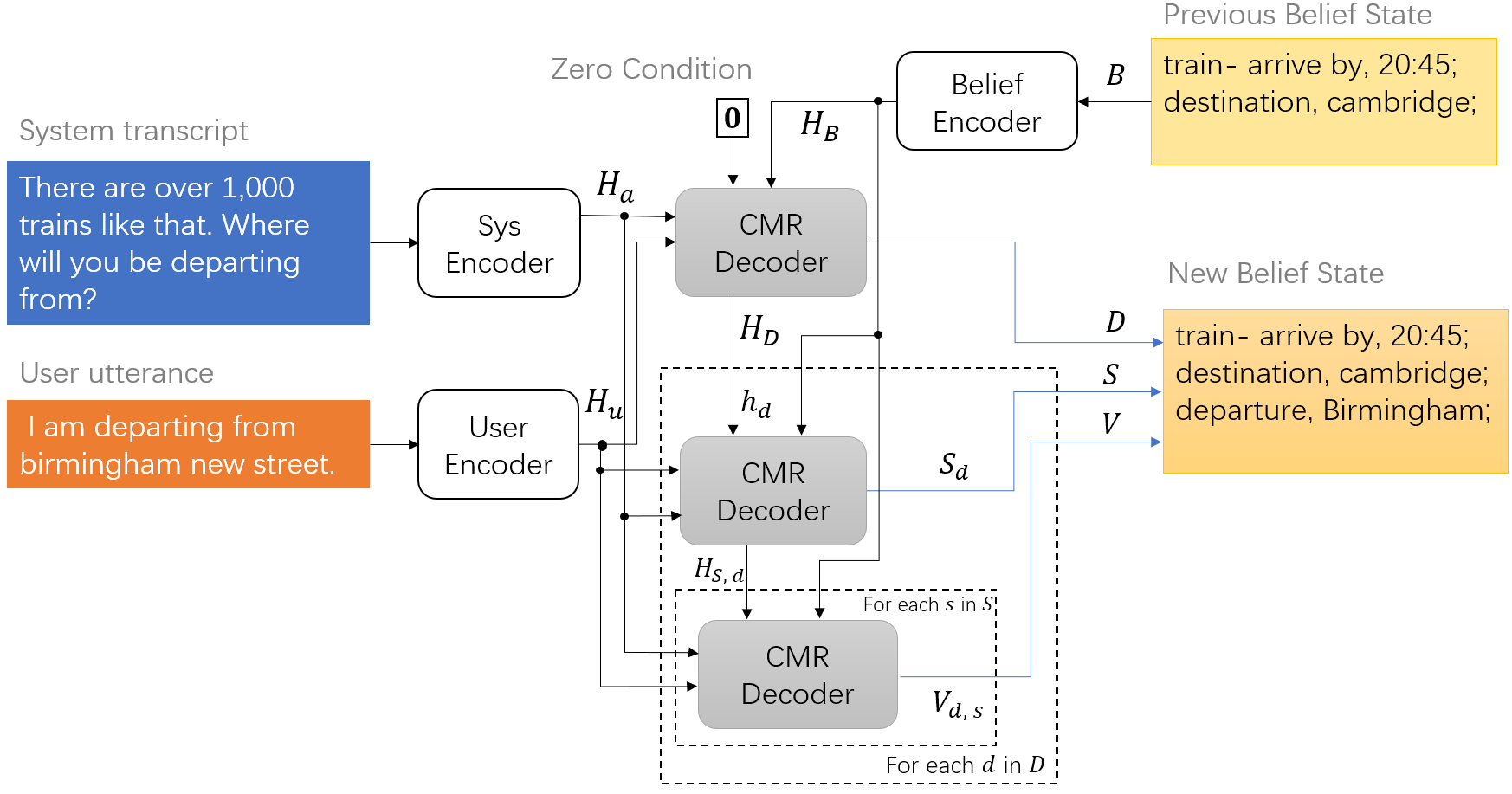}
\caption{The general model architecture of the Hierarchical Sequence Generation Network. The Conditional Memory Relation (CMR) decoders 
(gray) share all of their parameters.}\label{f3}
\end{figure*}



  

\section{Hierarchical Sequence Generation for DST}

Given a dialogue $D$ which consists of $T$ turns of user utterances and system actions, our target is to predict the state at each turn. Different from previous
methods
which formulate multi-label state prediction as a collection of binary prediction problems, COMER adapts the task into a sequence generation problem via a Seq2Seq framework. 

As
shown in \Cref{f3}, COMER consists of three encoders
and three hierarchically stacked decoders. We 
propose
a novel Conditional Memory Relation Decoder (CMRD) for sequence decoding. Each encoder includes an embedding layer and a BiLSTM. The encoders take in the user utterance, the 
previous system actions, and the previous belief states at the current turn, and encodes them into the embedding space. The user encoder and the system encoder use the fixed BERT model as the embedding layer.

Since the \emph{slot value} pairs are un-ordered set elements of a domain in the belief states, we first order the sequence of domain
according to their frequencies 
as they appear
in the training set \cite{yang2018sgm}, and then order the \emph{slot value} pairs in the domain according to the slot's frequencies of as they appear in a domain. After the sorting of the state elements, We represent the belief states following the paradigm: (Domain1- Slot1, Value1; Slot2, Value2; ... Domain2- Slot1, Value1; ...) for a more concise representation compared with the nested tuple representation.

All the CMRDs take the same representations from the system encoder, user encoder and the belief encoder as part of the input. In the procedure of hierarchical sequence generation, the first CMRD takes 
a
zero vector for its condition input $\mathbf{c}$, and generates a sequence of the domains, $D$, as well as the hidden representation of domains $H_D$. For each $d$ in $D$, the second CMRD then takes the corresponding $h_d$ as the condition input and generates the slot sequence $S_d$, and representations, $H_{S,d}$. Then for each $s$ in $S$, the third CMRD generates the value sequence $V_{d,s}$ based on the corresponding $h_{s,d}$. We update the belief state with the new $(d,(s_d,V_{d,s}))$ pairs and 
perform
the procedure
iteratively until a dialogue is completed. 
All the CMR decoders share all of their parameters.

Since our model generates domains and slots instead of taking pre-defined slots as 
inputs, and the number of 
domains and slots generated each turn is only related to the complexity of the contents 
covered 
in
a specific dialogue,
the inference time complexity of COMER is
$O(1)$ with respect to the number of pre-defined slots and values.

\subsection{Encoding Module}

Let $X$ represent a user utterance or system transcript consisting of a sequence of words $\{w_1,\ldots,w_T\}$. The encoder first passes the sequence $\{\mathit{[CLS]},w_1,\ldots,w_T,\mathit{[SEP]}\}$ into a pre-trained BERT model and obtains its contextual embeddings $E_{X}$. Specifically, we leverage the output of all layers of BERT and take the average to obtain the contextual embeddings.

For each domain/slot appeared in the training set, if it has more than one word, such as `price range', `leave at', etc., we feed it into BERT and take the average of the word vectors to form the extra slot embedding $E_{s}$.  In this way, we map each domain/slot to a static embedding, which allows us to generate a domain/slot as a whole instead of a token at each time step of domain/slot sequence decoding. We also construct a static vocabulary embedding $E_{v}$ by feeding each token in the BERT vocabulary into BERT. The final static word embedding $E$ is the concatenation of the $E_{v}$ and $E_{s}$.
  


After we obtain the contextual embeddings for the user utterance, system action, and the static embeddings for the previous belief state, we feed each of them 
into a Bidirectional LSTM \cite{lstm}. 
\begin{equation}
    \begin{aligned}
    \mathbf{h}_{a_t} & = \textrm{BiLSTM}(\mathbf{e}_{X_{a_t}}, \mathbf{h}_{a_{t-1}}) \\
    \mathbf{h}_{u_t} & = \textrm{BiLSTM}(\mathbf{e}_{X_{u_t}}, \mathbf{h}_{u_{t-1}}) \\
    \mathbf{h}_{b_t} & = \textrm{BiLSTM}(\mathbf{e}_{X_{b_t}}, \mathbf{h}_{b_{t-1}}) \\
    \mathbf{h}_{a_0} & = \mathbf{h}_{u_0} = \mathbf{h}_{b_0} = c_{0}, \\
    \end{aligned}
\end{equation}
where $c_{0}$ is the zero-initialized hidden state for the BiLSTM. The hidden size of the BiLSTM is $d_m/2$. We concatenate the forward and the backward hidden representations of each token from the BiLSTM to obtain the token representation $\mathbf{h}_{k_t}\in R^{d_m}$, $k\in\{a,u,b\}$ at each time step $t$. The hidden states of all time steps are concatenated to obtain the final representation of $H_{k}\in R^{T \times d_m}, k \in \{a,u,B\}$. The parameters are shared between all of the BiLSTMs.


\subsection{Conditional Memory Relation Decoder}

\begin{figure*}[htb]
\centering
\includegraphics[width=16cm]{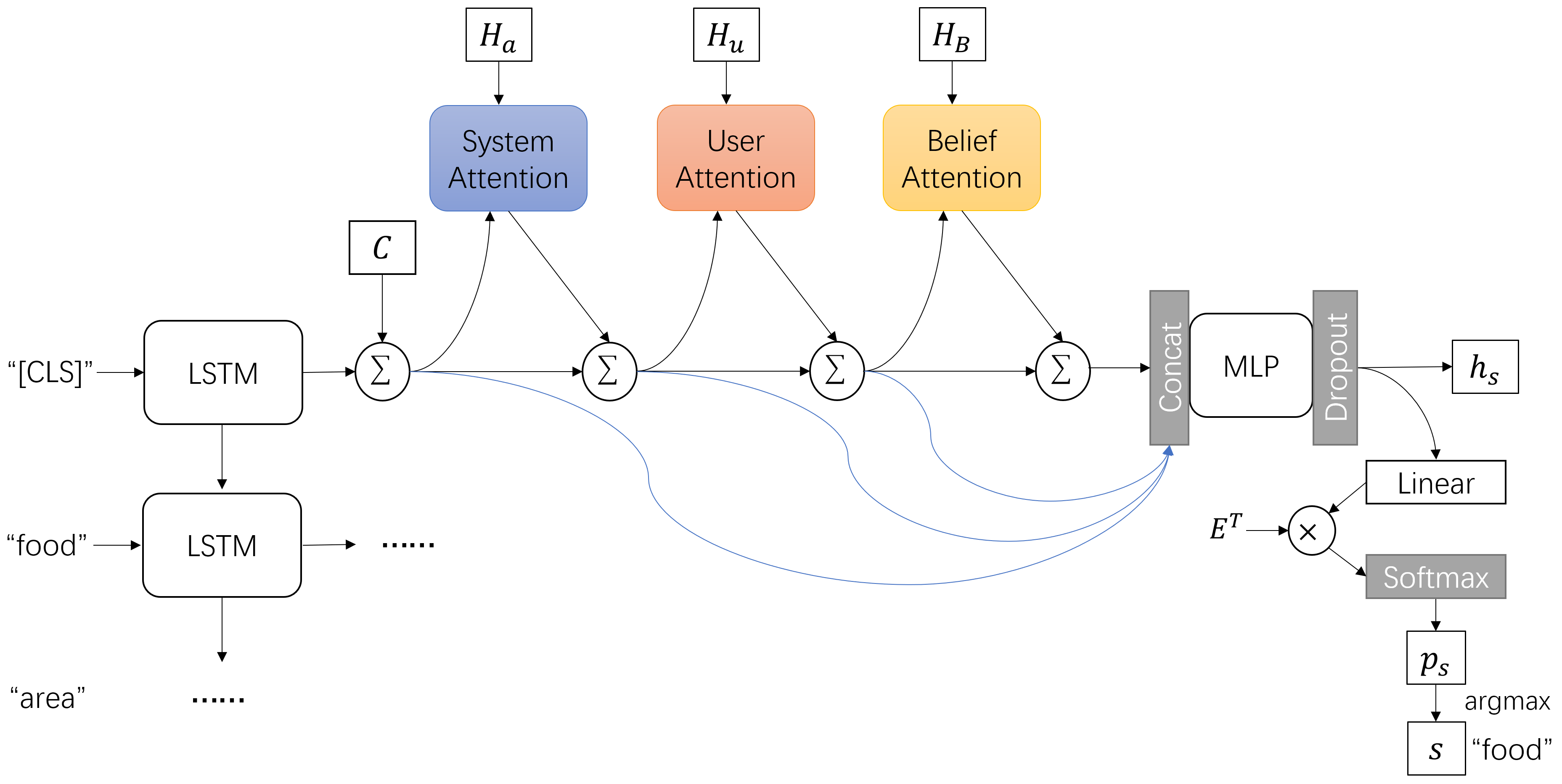}
\caption{The general structure of the Conditional Memory Relation Decoder. The decoder output, $s$ (e.g.~``food''), is refilled to the LSTM for 
the decoding of the next step. The blue lines in the figure means that the gradients are blocked during the back propagation stage.
}\label{f4}
\end{figure*}

Inspired by Residual Dense Networks \cite{rdb}, End-to-End Memory Networks \cite{memNN} and Relation Networks \cite{rn}, we here propose the Conditional Memory Relation Decoder (CMRD). Given a token embedding, $\mathbf{e}_x$, CMRD outputs the next token, $s$, and the hidden representation, $h_s$, with the hierarchical memory access of different encoded information sources, $H_B$, $H_a$, $H_u$, and the relation reasoning under a certain given condition $\mathbf{c}$, 
\[
    \mathbf{s}, \mathbf{h}_s= \textrm{CMRD}(\mathbf{e}_x, \mathbf{c}, H_B, H_a, H_u),
\]
the final output matrices $S,H_s \in R^{l_s\times d_m}$ are concatenations of all generated $\mathbf{s}$ and $\mathbf{h}_s$ (respectively) 
along the sequence length dimension,
where $d_m$ is the model size, and $l_s$ is the generated sequence length. The general structure of the CMR decoder is shown in Figure~\ref{f4}. Note that the CMR decoder can support 
additional
memory sources by adding the residual connection and the attention block, but here we only show the structure with three sources: belief state representation ($H_B$), system transcript representation ($H_a$), and user utterance representation ($H_u$), 
corresponding
to a
dialogue state tracking
scenario. Since we share the parameters between all of the decoders, thus CMRD is actually a \emph{2-dimensional} auto-regressive model with respect to both the condition generation and the sequence generation task.

At each time step $t$, the CMR decoder first embeds the token $x_t$ with a fixed token embedding $E\in R^{d_e\times d_v}$, where $d_e$ is the embedding size and $d_v$ is the vocabulary size. The initial token $x_0$ is ``[CLS]". The embedded vector $\textbf{e}_{x_t}$ is then encoded with an LSTM, which emits a hidden representation $\textbf{h}_0 \in R^{d_m}$,

\[
   \textbf{h}_0= \textrm{LSTM}(\textbf{e}_{x_t},\textbf{q}_{t-1}).
\]
where $\textbf{q}_t$ is the hidden state of the LSTM. $\textbf{q}_0$ is initialized with an average of the hidden states of the belief encoder, the system encoder and the user encoder which produces $H_B$, $H_a$, $H_u$ respectively. 

$\mathbf{h}_0$ is then summed (element-wise) with the condition representation $\mathbf{c}\in R^{d_m}$ to produce $\mathbf{h}_1$, which is (1) fed into the attention module; (2) used for residual connection; and (3) concatenated with other $\mathbf{h}_i$, ($i>1$) to produce the concatenated working memory, $\mathbf{r_0}$, for relation reasoning,
\begin{align*}
   \mathbf{h}_1 & =\mathbf{h}_0+\mathbf{c},\\
   \mathbf{h}_2 & =\mathbf{h}_1+\text{Attn}_{\text{belief}}(\mathbf{h}_1,H_e),\\
   \mathbf{h}_3 & = \mathbf{h}_2+\text{Attn}_{\text{sys}}(\mathbf{h}_2,H_a),\\
   \mathbf{h}_4 & = \mathbf{h}_3+\text{Attn}_{\text{usr}}(\mathbf{h}_3,H_u),\\ 
   \mathbf{r} & = \mathbf{h}_1\oplus \mathbf{h}_2\oplus\mathbf{h}_3\oplus\mathbf{h}_4 \in R^{4d_m},
\end{align*}
where $\text{Attn}_k$ ($k\in\{ \text{belief}, \text{sys},\text{usr}\}$) are the attention modules applied respectively to $H_B$, $H_a$, $H_u$, and $\oplus$ means the concatenation operator. The gradients are blocked for $ \mathbf{h}_1,\mathbf{h}_2,\mathbf{h}_3$ during the back-propagation stage, since we only need them to work as the supplementary memories for the relation reasoning followed.

The attention module takes a vector, $\mathbf{h}\in R^{d_m}$, and a matrix, $H\in R^{d_m\times l}$ as 
input, where $l$ is the sequence length of the representation, and outputs $\mathbf{h}_a$, a weighted sum of the column vectors in $H$.
\begin{align*}
    \mathbf{a} & =W_1^T\mathbf{h}+\mathbf{b}_1& &\in R^{d_m},\\
    \mathbf{c} &=\text{softmax}(H^Ta)& &\in R^l,\\
    \mathbf{h} &=H\mathbf{c}& &\in R^{d_m},\\
    \mathbf{h}_a &=W_2^T\mathbf{h}+\mathbf{b}_2& &\in R^{d_m},
\end{align*}
where the weights $W_1\in R^{d_m \times d_m}$, $W_2\in R^{d_m \times d_m}$ and the bias $b_1\in R^{d_m}$, $b_2\in R^{d_m}$ are the learnable parameters.

The order of the attention modules, \emph{i.e.,}~first attend to the system and the user and then the belief, is decided empirically. We can interpret this hierarchical structure as the internal order for the memory processing, since from the daily life experience, people tend to attend to the most contemporary memories (system/user utterance) first and then attend to the older history (belief states). All of the parameters are shared between the attention modules.

The concatenated working memory, $\mathbf{r}_0$, is then fed into a Multi-Layer Perceptron (MLP) with four layers,
\begin{align*}
   \mathbf{r}_1 & =\sigma(W_1^T\mathbf{r}_0+\mathbf{b}_1),\\
   \mathbf{r}_2 & =\sigma(W_2^T\mathbf{r}_1+\mathbf{b}_2),\\
   \mathbf{r}_3 & = \sigma(W_3^T\mathbf{r}_2+\mathbf{b}_3),\\
   \mathbf{h}_k & = \sigma(W_4^T\mathbf{r}_3+\mathbf{b}_4),
\end{align*}
where $\sigma$ is 
a
non-linear activation, and the weights $W_1 \in R^{4d_m \times d_m}$, $W_i \in R^{d_m \times d_m}$ and the bias $b_1 \in R^{d_m}$, $b_i \in R^{d_m}$ are
learnable parameters, and $2\leq i\leq 4$. The number of 
layers for 
the
MLP is decided by the grid search.

The hidden representation of the next token, $\mathbf{h}_k$, is first fed into a dropout layer with 
drop rate
$p$, and then (1) emitted out of the decoder as a representation; and (2) fed into a linear layer to generate the next token,
\begin{align*}
   \mathbf{h}_s & =\text{dropout}(\mathbf{h}_k)& &\in R^{d_m},\\
   \mathbf{h}_o & =W_k^T\mathbf{h}_s+\mathbf{b}_s& &\in R^{d_e},\\
    \mathbf{p}_s & =\text{softmax}(E^T\mathbf{h}_o)& &\in R^{d_v},\\
    s & =\text{argmax}(\mathbf{p}_s)& &\in R,  
\end{align*}
where the weight $W_k\in R^{d_m \times d_e}$ and the bias $b_k\in R^{d_e}$ are
learnable parameters. Since $d_e$ is the embedding size and the model parameters are independent of the vocabulary size, the CMR decoder can make predictions on a dynamic vocabulary and implicitly supports the generation of 
unseen words. When training the model, we minimize the cross-entropy loss between the output probabilities, $\mathbf{p}_s$, and the given labels.


\section{Experiments}
\subsection{Experimental Setting}

\begin{table}[t!]
\begin{center}
\begin{tabular}{lccc}
\toprule \bf Metric & \bf WoZ2.0 & \bf MultiWoZ   \\ \midrule
Avg. \# turns, $t$&  7.45 &  13.68  \\
Avg. tokens~/~turn, $s$ &  11.24 &   13.18\\
Number of Slots, $n$ & 3  & 35\\
Number of Values, $m$ & 99 & 4510 \\
Training set size  & 600 & 8438 \\
Validation set size  & 200 & 1000 \\
Test set size  & 400 & 1000 \\
\bottomrule
\end{tabular}
\end{center}
\caption{\label{t2} The statistics of the WoZ2.0 and the MultiWoZ datasets.}
\end{table} 

We first test our model on the single domain dataset, WoZ2.0 \cite{wenN2N17}. It consists of 1,200 dialogues from the restaurant reservation domain with three pre-defined slots: \emph{food}, \emph{price range}, and \emph{area}. Since the \emph{name} slot rarely occurs in the dataset, it is not included in our experiments,
following
previous literature \cite{ren2018emnlp,perez2017dialog}. 
Our model is also tested on the multi-domain dataset, MultiWoZ \cite{mwoz}. It has a more complex ontology with 7 domains and 25 
predefined slots.
Since the \emph{combined slot-value} pairs representation of the belief states has to be applied for the model with $O(n)$ ITC, the total number of 
slots 
is
35. 
The statistics of these two datsets are shown in Table \ref{t2}.

Based on the statistics from these two datasets, we can calculate the theoretical
Inference Time Multiplier (ITM), $K$, as a metric of scalability. Given the inference time complexity, 
ITM
measures how many times a model will be slower when being transferred from the WoZ2.0 dataset, $d_1$, to the MultiWoZ dataset, $d_2$,
\[
    K= h(t)h(s)h(n)h(m)\\
\]
\[
    h(x)=\left\{
    \begin{array}{lcl}
    1 & &O(x)=O(1),\\
    \frac{x_{d_2}}{x_{d_1}}& & \text{otherwise},\\

    \end{array}
    \right.
\]
where $O(x)$ means the Inference Time Complexity (ITC) of the variable $x$. For 
a
model having an ITC of $O(1)$ with respect to the number of slots $n$, and values $m$, the ITM will be a multiplier of 2.15x, while for an ITC of $O(n)$, it will be a multiplier of 25.1, and 1,143 for $O(mn)$.

As a convention, the metric of joint goal accuracy is used to compare our model to
previous work. The joint goal accuracy regards the model making a successful belief state prediction if all of the slots and values predicted are exactly matched with the labels provided. Since our model also generate un-ordered domains and slots, we only need to require the predicted belief dictionary to be exactly matched with the labels during model evaluation. We also did a simple post-processing to remove the generated (DOMAIN, SLOT, VALUE) triplets which have empty elements.
In this work, the model with the highest joint accuracy on the validation set is evaluated on the test set for the test joint accuracy measurement.


\subsection{Implementation Details}
We use the $\text{BERT}_\text{large}$ model for both 
contextual and
static embedding generation. All
LSTMs in the model are stacked with 2 layers, and only the output of the last layer is taken as a hidden representation. 
ReLU non-linearity is used for the activation function, $\sigma$.

The hyper-parameters of our model are identical for both the WoZ2.0 and the MultiwoZ datasets: dropout rate $p=0.5$, model size $d_m=512$, embedding size $d_e=1024$. For 
training on
WoZ2.0, the model is trained with a batch size of 32 and the ADAM optimizer \cite{adam} for 150 epochs, while for 
MultiWoZ, the AMSGrad optimizer \cite{ams} and a batch size of 16 is adopted for 
15 epochs of training.
For both optimizers, we use a learning rate of 0.0005 with 
a
gradient clip of 2.0. We initialize all
weights in our model with
Kaiming initialization \cite{kaiming} and adopt zero initialization for the bias. All
experiments are conducted on a single NVIDIA GTX 1080Ti GPU.

\begin{table*}[t!]
\begin{center}
\begin{tabular}{lccc}
\toprule \bf DST Models & \begin{tabular}[c]{@{}c@{}}\bf Joint Acc.\\\bf WoZ 2.0\end{tabular}   &\begin{tabular}[c]{@{}c@{}}\bf Joint Acc.\\\bf MultiWoZ \end{tabular}       &\bf ITC \\ \midrule
Baselines \cite{mrkvsic2017neural}  &  70.8\%  & 25.83\% & $O(mn)$\\
NBT-CNN \cite{mrkvsic2017neural}& 84.2\% & -& $O(mn)$     \\
StateNet\_PSI \cite{ren2018emnlp}& \bf 88.9\% & - & $O(n)$   \\
GLAD \cite{nouri2018scalable}  & 88.5\% & 35.58\%  & $O(mn)$  \\
HyST (ensemble) \cite{HyST} & - &  44.22\% & \bf $O(n)$ \\
DSTRead (ensemble) \cite{dstread} & - &  42.12\% & \bf $O(n)$ \\
TRADE \cite{trade} & - &  48.62\% & \bf $O(n)$ \\
COMER & 88.6\% &  \bf 48.79\% & \bf $O(1)$ \\
\bottomrule
\end{tabular}
\end{center}
\caption{\label{c} The joint goal accuracy of the DST models on the WoZ2.0 test set and the MultiWoZ test set. We also include the Inference Time Complexity (ITC) for each model as a metric for scalability. The baseline accuracy for the WoZ2.0 dataset is the Delexicalisation-Based (DB) Model \cite{mrkvsic2017neural}, while the baseline for the MultiWoZ dataset is taken from the official website of MultiWoZ \cite{mwoz}.
}
\end{table*} 

\subsection{Results}

To measure the actual inference time multiplier of our model, we evaluate the runtime of the best-performing models on the validation sets of both the WoZ2.0 and 
MultiWoZ datasets. During evaluation, we set the batch size to 1 to avoid the influence of 
data parallelism and 
sequence padding. On the validation set of WoZ2.0, we 
obtain
a runtime of 65.6 seconds, while on 
MultiWoZ, the runtime is 835.2 seconds. 
Results are averaged 
across
5 runs. Considering 
that the validation set of MultiWoZ is 5 times larger than 
that of
WoZ2.0, the actual inference time multiplier is 2.54 for our model. Since the actual inference time multiplier 
roughly of
the same magnitude as
the theoretical value of 2.15, we can 
confirm empirically
that we have the $O(1)$ inference time complexity and
thus obtain
full scalability to the number of slots and values pre-defined in an ontology.

\Cref{c} compares our model with the previous state-of-the-art on both the WoZ2.0 test set and the MultiWoZ test set. For the WoZ2.0 dataset, we maintain 
performance at the level of the state-of-the-art, with a marginal drop of 0.3\% compared with 
previous work. Considering the fact that WoZ2.0 is a relatively small dataset, this 
small
difference
does not represent a significant
big performance 
drop.
On the muli-domain dataset, MultiWoZ, our model achieves a joint goal accuracy of 48.79\%, which marginally outperforms the previous state-of-the-art.

\subsection{Ablation Study}

\begin{table}[t!]
\begin{center}
\begin{tabular}{lc}
\toprule \bf Model & \bf Joint Acc.\\ \midrule
COMER  & 88.64\% \\
- Hierachical-Attn  & 86.69\% \\
- MLP  & 83.24\% \\
\bottomrule
\end{tabular}
\end{center}
\caption{\label{ab} The ablation study on the WoZ2.0 dataset with the joint goal accuracy on the test set. For ``- Hierachical-Attn", we remove the residual connections between the attention modules in the CMR decoders and all the attention memory access are based on the output from the LSTM. For ``- MLP", we further replace the MLP with a single linear layer with the non-linear activation. }
\end{table} 

\begin{table}[t!]
\begin{center}
\begin{tabular}{lccc}
\toprule \bf Model & \bf JD Acc. & \bf JDS Acc.& \bf JG Acc.\\ \midrule
COMER  &95.52\% &55.81\% & 48.79\% \\
- moveDrop &95.34\% &55.08\% & 47.19\% \\
- postprocess &95.53\% &54.74\% & 45.72\% \\
- ShareParam &94.96\% &54.40\% & 44.38\% \\
- Order  &95.55\% & 55.06\% & 42.84\% \\
- Nested  &- &49.58\% & 40.57\% \\
- BlockGrad &- & 49.36\% & 39.15\% \\
\bottomrule
\end{tabular}
\end{center}
\caption{\label{abm} The ablation study on the MultiWoZ dataset with the joint domain accuracy (JD Acc.), joint domain-slot accuracy (JDS Acc.) and joint goal accuracy (JG Acc.) on the test set. For ``- moveDrop'', we move the dropout layer to be in front of the final linear layer before the Softmax. For ``- postprocess'', we further fix the decoder embedding layer and remove the post-processing during model evaluation. For ``- ShareParam'', we further remove the parameter sharing mechanism on the encoders and the attention modules. For ``- Order'', we further arrange the order of the slots according to its global frequencies in the training set instead of the local frequencies given the domain it belongs to. For ``- Nested'', we do not generate domain sequences but generate \emph{combined slot} sequences which combines the domain and the slot together. For ``- BlockGrad'', we further remove the gradient blocking mechanism in the CMR decoder.}
\end{table} 

To prove the effectiveness of our 
structure 
of the Conditional Memory Relation Decoder (CMRD), 
we conduct
ablation experiments 
on the WoZ2.0 dataset.
The effectiveness of our hierarchical attention design is proved by an accuracy drop of 1.95\% after removing 
residual connections and the hierarchical stack of our attention modules. 
The accuracy drop of removing MLP is partly due to the reduction of
the
number of model parameters, but it also proves that stacking more layers in an MLP can improve the relational reasoning performance given a concatenation of multiple representations from different sources.

We also conduct the ablation study on the MultiWoZ dataset for a more precise analysis on the hierarchical generation process.  For joint domain accuracy, we calculate the probability that all domains generated in each turn are exactly matched with the labels provided. The joint domain-slot accuracy further calculate the probability that all domains and slots generated are correct, while the joint goal accuracy requires all the domains, slots and values generated are exactly matched with the labels. From \Cref{abm}, We can further calculate that given the correct slot prediction, COMER has 87.42\% chance to make the correct value prediction. While COMER has done great job on domain prediction (95.52\%) and value prediction (87.42\%), the accuracy of the slot prediction given the correct domain is only 58.43\%. We suspect that this is because we only use the previous belief state to represent the dialogue history, and the inter-turn reasoning ability on the slot prediction suffers from the limited context and the accuracy is harmed due to the multi-turn mapping problem \cite{trade}. We can also see that the JDS Acc. has an absolute boost of 5.48\% when we switch from the \emph{combined slot} representation to the nested tuple representation. This is because the subordinate relationship between the domains and the slots can be captured by the hierarchical sequence generation, while this relationship is missed when generating the domain and slot together via the \emph{combined slot} representation.

\subsection{Qualitative Analysis}

\begin{figure*}[htb]
\centering
\includegraphics[width=16cm]{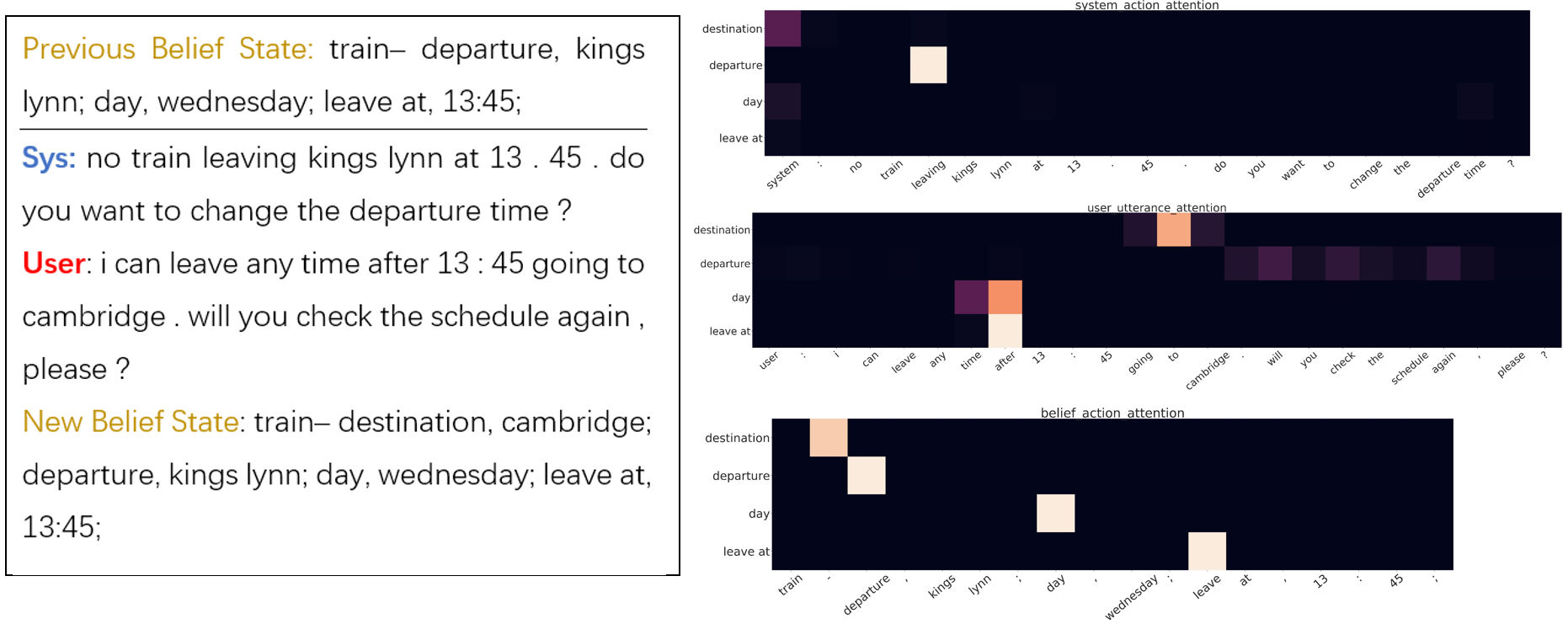}
\caption{An example belief prediction of our model on the MultiWoZ test set. The attention scores for belief states, system transcript and user utterance in CMRD is visualized 
on the right.
Each row corresponds to the attention score of the generation step of a particular slot under the `train' domain.}\label{f5}
\end{figure*}

\Cref{f5} shows an example of the belief state prediction result in one turn of a dialogue on the MultiWoZ test set. The visualization includes the CMRD attention scores over the belief states, system transcript and user utterance during the decoding stage of the slot sequence.

From the system attention (top right), since it is the first attention module and no previous context information is given, it can only find the information indicating the slot ``departure'' from the system utterance under the domain condition, and attend to the evidence ``leaving'' correctly during the generation step of ``departure''. From the user attention, we can see that it captures the most helpful keywords that 
are
necessary for correct prediction, such as ``after" for ``day" and ``leave at'', ``to" for ``destination". 
Moreover, during the generation step of ``departure'', the user attention successfully discerns that, based on the context, the word ``leave'' is not the evidence that need to be accumulated and choose to attend nothing in this step.
For the belief attention, 
we can see that the belief attention module correctly attends to a previous slot for each generation step of a slot that has been presented in the previous state.  For the generation step of the new slot ``destination", since the previous state does not have the ``destination" slot, the belief attention module only attends to the `-' mark after the `train' domain to indicate that the generated word should belong to this domain.


\section{Related Work}

\noindent  \textbf{Semi-scalable Belief Tracker}
\citet{rastogi2017scalable} proposed an approach that can generate fixed-length candidate sets for each of the slots from the dialogue history. Although they only need to 
perform
inference for a fixed number of 
values, they still need to 
iterate
over all 
slots defined in the ontology to make a prediction for a given dialogue turn. 
In addition,
their method needs an external language understanding module to extract the exact entities from a dialogue to form 
candidates, which will not work if the label value is an abstraction and does not have the exact match with the words in the dialogue.

StateNet \cite{ren2018emnlp} achieves
state-of-the-art performance with the property that its parameters are independent of the number of slot values in the candidate set, and it also supports online training or inference with dynamically changing slots and values. Given a slot that
needs
tracking, it only needs to 
perform inference once to
make
the prediction for a turn, but this also means that its inference time complexity 
is
proportional to the number of slots. 

TRADE \cite{trade} achieves state-of-the-art performance on the MultiWoZ dataset by applying the copy mechanism for the value sequence generation.
Since TRADE takes $n$ combinations of the domains and slots as the input, the inference time complexity of TRADE is $O(n)$. The performance improvement achieved by TRADE is mainly due to the fact that it incorporates the copy mechanism that can boost the accuracy on the ‘name’ slot, which mainly needs the ability in copying names from the dialogue history. However, TRADE does not report its performance on the WoZ2.0 dataset which does not have the ‘name’ slot. 

DSTRead \cite{dstread} formulate the dialogue state tracking task as a reading comprehension problem by asking slot specified questions to the BERT model and find the answer span in the dialogue history for each of the pre-defined \emph{combined slot}. Thus its inference time complexity is still $O(n)$. This method suffers from the fact that its generation vocabulary is limited to the words occurred in the dialogue history, and it has to do a manual combination strategy with another joint state tracking model on the development set to achieve better performance.

\noindent  \textbf{Contextualized Word Embedding (CWE)} was first proposed by \citet{elmo}. Based on 
the intuition
that the meaning of a word is highly correlated with its context, CWE takes the 
complete
context (sentences, passages, etc.) as the input, and outputs the corresponding word vectors that are unique under the given context. Recently, with the success of 
language models (e.g.~\citet{bert}) that are trained on 
large scale data, 
contextualizeds word embedding 
have been
further improved and can achieve the same 
performance compared to 
(less flexible) 
finely-tuned
pipelines.

\noindent \textbf{Sequence Generation Models.} Recently, sequence generation models have been successfully applied in the realm of
multi-label classification (MLC) \citep{yang2018sgm}.
Different from traditional binary relevance methods, they proposed a sequence generation model for MLC tasks which takes into consideration the correlations between labels. Specifically, the model follows the encoder-decoder structure with an attention mechanism \citep{Cho2014encoder}, where the decoder generates a sequence of labels. 
Similar to language modeling tasks, the decoder output at each time step will be conditioned on the previous predictions during  generation. Therefore the correlation between generated labels 
is
captured by the decoder.

\section{Conclusion}

In this work, we proposed the Conditional Memory Relation Network (COMER), the first dialogue state tracking model that has a constant inference time complexity with respect to the number of domains,
slots and values pre-defined in an ontology. Besides 
its
scalability, the joint goal accuracy of our model also achieve the similar performance compared with the state-of-the-arts on both the MultiWoZ dataset and the WoZ dataset. Due to the flexibility of our hierarchical encoder-decoder framework and the CMR decoder, abundant future research direction remains as applying the transformer structure, incorporating open vocabulary and copy mechanism for explicit unseen words generation, and inventing better dialogue history access mechanism to accommodate efficient inter-turn reasoning.


\noindent\textbf{Acknowledgements.} This work is partly supported by NSF \#1750063. We thank all the reviewers for their constructive suggestions. We also want to thank Zhuowen Tu and Shengnan Zhang for the early discussions of the project.

\bibliography{emnlp-ijcnlp-2019}

\begin{thebibliography}{27}
\expandafter\ifx\csname natexlab\endcsname\relax\def\natexlab#1{#1}\fi

\bibitem[{Budzianowski et~al.(2018)Budzianowski, Wen, Tseng, Casanueva, Ultes,
  Ramadan, and Gasic}]{mwoz}
Pawel Budzianowski, Tsung-Hsien Wen, Bo-Hsiang Tseng, I{\~n}igo Casanueva,
  Stefan Ultes, Osman Ramadan, and Milica Gasic. 2018.
\newblock Multiwoz - a large-scale multi-domain wizard-of-oz dataset for
  task-oriented dialogue modelling.
\newblock In \emph{EMNLP}.

\bibitem[{Cho et~al.(2014)Cho, van Merrienboer, aglar Gülehre, Bahdanau,
  Bougares, Schwenk, and Bengio}]{Cho2014encoder}
Kyunghyun Cho, Bart van Merrienboer, aglar Gülehre, Dzmitry Bahdanau, Fethi
  Bougares, Holger Schwenk, and Yoshua Bengio. 2014.
\newblock Learning phrase representations using rnn encoder-decoder for
  statistical machine translation.
\newblock In \emph{EMNLP}.

\bibitem[{Devlin et~al.(2018)Devlin, Chang, Lee, and Toutanova}]{bert}
Jacob Devlin, Ming-Wei Chang, Kenton Lee, and Kristina Toutanova. 2018.
\newblock Bert: Pre-training of deep bidirectional transformers for language
  understanding.
\newblock In \emph{NAACL-HLT}.

\bibitem[{Gao et~al.(2019)Gao, Sethi, Agarwal, Chung, and Hakkani}]{dstread}
Shuyang Gao, Abhishek Sethi, Sanchit Agarwal, Tagyoung Chung, and Dilek~Zeynep
  Hakkani. 2019.
\newblock Dialog state tracking: A neural reading comprehension approach.
\newblock \emph{ArXiv}, abs/1908.01946.

\bibitem[{Goel et~al.(2019)Goel, Paul, and Hakkani}]{HyST}
Rahul Goel, Shachi Paul, and Dilek~Zeynep Hakkani. 2019.
\newblock Hyst: A hybrid approach for flexible and accurate dialogue state
  tracking.
\newblock \emph{ArXiv}, abs/1907.00883.

\bibitem[{He et~al.(2015)He, Zhang, Ren, and Sun}]{kaiming}
Kaiming He, Xiangyu Zhang, Shaoqing Ren, and Jian Sun. 2015.
\newblock Delving deep into rectifiers: Surpassing human-level performance on
  imagenet classification.
\newblock \emph{2015 IEEE International Conference on Computer Vision (ICCV)},
  pages 1026--1034.

\bibitem[{Henderson et~al.(2014)Henderson, Thomson, and
  Williams}]{henderson-thomson-williams:2014:W14-43}
Matthew Henderson, Blaise Thomson, and Jason~D. Williams. 2014.
\newblock The second dialog state tracking challenge.
\newblock In \emph{SIGDIAL Conference}.

\bibitem[{Hochreiter and Schmidhuber(1997)}]{lstm}
Sepp Hochreiter and J{\"u}rgen Schmidhuber. 1997.
\newblock Long short-term memory.
\newblock \emph{Neural Computation}, 9:1735--1780.

\bibitem[{Kingma and Ba(2015)}]{adam}
Diederik~P. Kingma and Jimmy Ba. 2015.
\newblock Adam: A method for stochastic optimization.
\newblock \emph{CoRR}.

\bibitem[{Liu and Perez(2017)}]{perez2017dialog}
Fei Liu and Julien Perez. 2017.
\newblock Dialog state tracking, a machine reading approach using memory
  network.
\newblock In \emph{EACL}.

\bibitem[{Mrksic et~al.(2017)Mrksic, S{\'e}aghdha, Wen, Thomson, and
  Young}]{mrkvsic2017neural}
Nikola Mrksic, Diarmuid~{\'O} S{\'e}aghdha, Tsung-Hsien Wen, Blaise Thomson,
  and Steve~J. Young. 2017.
\newblock Neural belief tracker: Data-driven dialogue state tracking.
\newblock In \emph{ACL}.

\bibitem[{Nouri and Hosseini-Asl(2018)}]{nouri2018scalable}
Elnaz Nouri and Ehsan Hosseini-Asl. 2018.
\newblock Toward scalable neural dialogue state tracking model.
\newblock \emph{arXiv preprint arXiv:1812.00899}.

\bibitem[{Peters et~al.(2018)Peters, Neumann, Iyyer, Gardner, Clark, Lee, and
  Zettlemoyer}]{elmo}
Matthew~E. Peters, Mark Neumann, Mohit Iyyer, Matt Gardner, Christopher Clark,
  Kenton Lee, and Luke~S. Zettlemoyer. 2018.
\newblock Deep contextualized word representations.
\newblock In \emph{NAACL-HLT}.

\bibitem[{Ramadan et~al.(2018)Ramadan, Budzianowski, and
  Gasic}]{ramadan2018large}
Osman Ramadan, Pawel Budzianowski, and Milica Gasic. 2018.
\newblock Large-scale multi-domain belief tracking with knowledge sharing.
\newblock In \emph{ACL}.

\bibitem[{Rastogi et~al.(2017)Rastogi, Hakkani-Tur, and
  Heck}]{rastogi2017scalable}
Abhinav Rastogi, Dilek Hakkani-Tur, and Larry Heck. 2017.
\newblock Scalable multi-domain dialogue state tracking.
\newblock \emph{arXiv preprint arXiv:1712.10224}.

\bibitem[{Reddi et~al.(2018)Reddi, Kale, and Kumar}]{ams}
Sashank~J. Reddi, Satyen Kale, and Sanjiv Kumar. 2018.
\newblock On the convergence of adam and beyond.
\newblock In \emph{ICLR}.

\bibitem[{Ren et~al.(2018)Ren, Xie, Chen, and Yu}]{ren2018emnlp}
Liliang Ren, Kaige Xie, Lu~Chen, and Kai Yu. 2018.
\newblock Towards universal dialogue state tracking.
\newblock In \emph{EMNLP}.

\bibitem[{Santoro et~al.(2017)Santoro, Raposo, Barrett, Malinowski, Pascanu,
  Battaglia, and Lillicrap}]{rn}
Adam Santoro, David Raposo, David G.~T. Barrett, Mateusz Malinowski, Razvan
  Pascanu, Peter~W. Battaglia, and Timothy~P. Lillicrap. 2017.
\newblock A simple neural network module for relational reasoning.
\newblock In \emph{NIPS}.

\bibitem[{Sennrich et~al.(2016)Sennrich, Haddow, and Birch}]{bpe}
Rico Sennrich, Barry Haddow, and Alexandra Birch. 2016.
\newblock Neural machine translation of rare words with subword units.
\newblock \emph{CoRR}, abs/1508.07909.

\bibitem[{Sukhbaatar et~al.(2015)Sukhbaatar, Szlam, Weston, and Fergus}]{memNN}
Sainbayar Sukhbaatar, Arthur Szlam, Jason Weston, and Rob Fergus. 2015.
\newblock End-to-end memory networks.
\newblock In \emph{NIPS}.

\bibitem[{Wen et~al.(2017)Wen, Gasic, Mrksic, Rojas-Barahona, hao Su, Ultes,
  Vandyke, and Young}]{wenN2N17}
Tsung-Hsien Wen, Milica Gasic, Nikola Mrksic, Lina~Maria Rojas-Barahona, Pei
  hao Su, Stefan Ultes, David Vandyke, and Steve~J. Young. 2017.
\newblock A network-based end-to-end trainable task-oriented dialogue system.
\newblock In \emph{EACL}.

\bibitem[{Wu et~al.(2019)Wu, Madotto, Hosseini-Asl, Xiong, Socher, and
  Fung}]{trade}
Chien-Sheng Wu, Andrea Madotto, Ehsan Hosseini-Asl, Caiming Xiong, Richard
  Socher, and Pascale Fung. 2019.
\newblock Transferable multi-domain state generator for task-oriented dialogue
  systems.
\newblock In \emph{ACL}.

\bibitem[{Xu and Hu(2018)}]{xu2018acl}
Puyang Xu and Qi~Hu. 2018.
\newblock An end-to-end approach for handling unknown slot values in dialogue
  state tracking.
\newblock In \emph{ACL}.

\bibitem[{Yang et~al.(2018)Yang, Sun, Li, Ma, Wu, and Wang}]{yang2018sgm}
Pengcheng Yang, Xu~Sun, Wei Li, Shuming Ma, Wei Wu, and Houfeng Wang. 2018.
\newblock Sgm: Sequence generation model for multi-label classification.
\newblock In \emph{COLING}.

\bibitem[{Young et~al.(2013)Young, Ga{\v{s}}i{\'c}, Thomson, and
  Williams}]{young2013pomdp}
Steve Young, Milica Ga{\v{s}}i{\'c}, Blaise Thomson, and Jason~D Williams.
  2013.
\newblock Pomdp-based statistical spoken dialog systems: A review.
\newblock \emph{Proceedings of the IEEE}, 101(5):1160--1179.

\bibitem[{Zhang et~al.(2018)Zhang, Tian, Kong, Zhong, and Fu}]{rdb}
Yulun Zhang, Yapeng Tian, Yu~Kong, Bineng Zhong, and Yun Fu. 2018.
\newblock Residual dense network for image super-resolution.
\newblock \emph{2018 IEEE/CVF Conference on Computer Vision and Pattern
  Recognition}, pages 2472--2481.

\bibitem[{Zhong et~al.(2018)Zhong, Xiong, and Socher}]{zhong2018global}
Victor Zhong, Caiming Xiong, and Richard Socher. 2018.
\newblock Global-locally self-attentive dialogue state tracker.
\newblock \emph{CoRR}, abs/1805.09655.

\end{thebibliography}
\bibliographystyle{acl_natbib}

\end{document}